\documentclass[runningheads]{llncs}
\usepackage[T1]{fontenc}
\usepackage{graphicx}
\begin{document}
\title{Empowering Domain-Specific Language Models with Graph-Oriented Databases: A Paradigm Shift in Performance and Model Maintenance}
\titlerunning{Empowering Domain-Specific Language Models with Graph-Oriented DB}
%

\author{Ricardo Di Pasquale\inst{1,2}  
\and \\
Soledad Represa\inst{3} }
%
%

\institute{Pontificia Universidad Catolica Argentina, Facultad de Ingenieria y Ciencias Agrarias \\ 
\email{rdipasquale@uca.edu.ar}\\ 
\and Acccenture \\
\email{ricardo.di.pasquale@accenture.com}\\ 
\and 
Pontificia Universidad Catolica Argentina, Facultad de Ingenieria y Ciencias Agrarias, Laboratorio de Ciencias de Datos e Inteligencia Artificial \\
\email{natacharepresa@uca.edu.ar}}

\maketitle              
\begin{abstract}
In an era dominated by data, the management and utilization of domain-specific language have emerged as critical challenges in various application domains, particularly those with industry-specific requirements. Our work is driven by the need to effectively manage and process large volumes of short text documents inherent in specific application domains. By leveraging domain-specific knowledge and expertise, our approach aims to shape factual data within these domains, thereby facilitating enhanced utilization and understanding by end-users. Central to our methodology is the integration of domain-specific language models with graph-oriented databases, facilitating seamless processing, analysis, and utilization of textual data within targeted domains. Our work underscores the transformative potential of the partnership of domain-specific language models and graph-oriented databases. This cooperation aims to assist researchers and engineers in metric usage, mitigation of latency issues, boosting explainability, enhancing debug and improving overall model performance.  Moving forward, we envision our work as a guide AI engineers, providing valuable insights for the implementation of domain-specific language models in conjunction with graph-oriented databases, and additionally provide valuable experience in full-life cycle maintenance of this kind of products.

\end{abstract}
\section{Introduction}
In recent years, the field of artificial intelligence has witnessed a paradigm shift with the advent of Large Language Models (LLMs). These advanced models, capable of processing and generating human-like text at an unprecedented scale, have revolutionized various domains, including natural language processing, machine learning, and information retrieval.

The integration of LLMs into practical applications across various industrial sectors has taken distinct paths. What follows is not an exhaustive list, but it is an effort to summarize the principal implementation styles: $(1)$ building a model from scratch, $(2)$ using a foundational model as it is, $(3)$ fine-tuning a foundational model, $(4)$ prompt engineering and retrieval augmented generation. 

From our own experience in Technology Consulting Services in the field of industry-specific solutions, i.e. domain-specific large language model applications, solutions of type $(3)$ and $(4)$  have been chosen more frequently. This kind of domain-specific applications have to deal with many key aspects such as: specific language (jargon or argot), complex but specific behaviour rules that are not part of general scope model training, complex maintenance due to changes in domain-specific rules, among others.

Modern GODB have emerged as a solution for highly-connected data, and link oriented queries and algorithms \cite{graphdb}. In fact, they have been a valuable solution in software industry for decades. The implementation of GODB in business solutions is not restricted to specific domains, and it's usually well suited for on-line transaction processing (OLTP) solutions as well as analytics (machine learning, business intelligence, data mining, etc.) solutions \cite{bechberger}.

In 2023 and 2024, numerous studies have underscored the significance of integrating knowledge graphs (KG) with Large Language Models (LLMs) in artificial intelligence (AI) \cite{jovanovic}\cite{pan}. This work aims to illustrate the value of the partnership between GODB and LLMs in industry-specific applications where extensive document analysis is required. The structure of this work is as follows: first, we will delineate the types of solutions under examination in section 2. In section 3, we will define KGs and explore their automated creation in GODB using LLMs. Section 4 will delve into Retrieval-Augmented Generation techniques with GODB support. In section 5, we will explore explainability techniques. Memory, context, and personalization will be addressed in section 6. Section 7 will highlight GODB's role in enabling factuality checks to mitigate the hallucination effect. Finally, we will conclude with a section on future directions.

\section{Domain-specific LLM solutions for industry-specific document processing and querying}
Generative AI and LLMs are supposed to have a huge impact transforming work across industries \cite{accen}. Domain-specific (DS) LLM solutions are well suited for industry-specific applications for many reasons. Among them, we can mention $(1)$ specialized vocabulary: this kind of applications need to apply a specific jargon and technical terms for dealing with  complex industry processes which were not included in original foundational model training. Usually text/audio/videos ingested in these solutions make extensive use of DS language. DS LLMs are trained or tuned on data from the particular industry, enabling them to better understand and generate contextually appropriate text using the specialized vocabulary of that domain. $(2)$ DS rules need to be used to deal with data, extract knowledge and interact with this data. $(3)$ Higher Accuracy: by focusing on a specific domain, LLMs can achieve higher accuracy in understanding and generating text related to that domain. These models are enabled to learn patterns and nuances more effectively. $(4)$ Improved Relevance: DS LLMs can provide more relevant responses to queries or prompts within their designated industry. They understand the context better and can produce text that aligns with the specific requirements and expectations of users within that industry. $(5)$ Better performance on niche tasks. $(6)$ Faster Deployment and Adoption. Since DS LLMs are tailored to meet the needs of a particular industry, they often require less fine-tuning and customization to be deployed effectively. $(7)$ Compliance and Security, industries have strict compliance and security requirements. DS LLMs can be tuned to adhere to these regulations and ensure that generated text meets industry standards for privacy, security, and legal compliance.

Common LLM use cases in industry-specific with large number of text data to be analyzed include: $(1)$ financial sensitivity analytics from day-to-day operations, $(2)$ concurrent comparison of commercial contracts and question-and-answer (Q\&A) sessions, $(3)$ spend classification, $(4)$ trend sensing in social networks, $(5)$ analysis of meeting minute and many others. For illustrative purposes, let's delve into the technical details of one such case: a meeting minute analyzer. Consider a scenario where we have a construction materials company (comprising both a factory and distribution network). Imagine a sizable team of sales representatives who visit clients for sales purposes while also addressing customer satisfaction, assessing materials quality, and gathering feedback on logistics and the industrial process. After each meeting, these representatives submit a brief note (potentially in audio format) to the central office, probably using technical and colloquial jargon. These notes are unstructured and diverse, possibly containing sentiment analysis, customer complaints about material quality, feedback on time-to-market, pricing comparisons with competitors, and more. Furthermore, suppose there are hundreds of sales representatives, each averaging five client visits per day, resulting in approximately half a million notes annually. The company aims to develop a software solution capable of ingesting this data, performing advanced analytics and generating insights. These insights will then be integrated into a chatbot or Q\&A agent, enabling real-time access to complex business information. From now on, we will refer to this case as \textit{"our case"}.

\section{Knowledge Graph}
The concept of KG has evolved over time \cite{bordes}, there isn't a single work in literature that definitively defined KG. Instead, the idea has emerged gradually through various contributions in the fields of AI, databases, semantic web, and information retrieval. In \cite{ehrlinger} authors provide a survey on potential deﬁnitions of KG. Since knowledge bases emerged more than a decade ago as fully functional products \cite{google}, we will take this source as the most important influence in knowledge graph definition. Let's then define a KG as a sub-type of Resource Description Framework (RDF) graph. An RDF graph is a directed graph that uses triples to describe web resources. An RDF graph \cite{ehrlinger}\cite{farber} consists of a set $R$ of RDF triples. For simplification, we should consider each RDF triple $(s, p, o)$ as a tuple of the following terms: $(1)$ a subject $s \in S$, $(2)$ a predicate $p \in P$, and $(3)$ an object $o \in O$. This defines a KG as a directed multigraph $G=(V,E)$ where $V=S \cup O$ and $E$ is the set of edges which derives from the links between elements of $S$ and $O$ derived from the triples defined in $R$. 

KGs and LLMs are part of an emerging field of cross-fertilization. Both can readily benefit from each other. Cooperation can take the form of KG-enhanced LLMs, LLM-augmented KGs, or synergized LLMs + KGs \cite{pan}. The advantage of implementing a KG on GODB is significant and self-explanatory. We won’t enumerate the characteristics of GODB, but we assert that most of them provide benefits to any KG implementation, such as query language, governance, scaling, etc.

\paragraph{Knowledge acquisition}
Knowledge acquisition \cite{peng} is central for our case: the \textbf{creation and updating of the KG}. LLMs can efficiently assist on this task by easily extracting entities and labeled relations between them. Graph-oriented databases (GODB) are necessary in our case to provide a sustainable environment for this large database. Additionally, most foundational models can handle specific GODB query languages, such as Cypher expressions for a Neo4j database. This process can be integrated into a data pipeline framework using data engineering techniques. There are key points AI engineer should take into account during this phase:

\paragraph{Prompt engineering during ingestion}
Prompt engineering techniques enable both the extraction of knowledge from documents and the construction of the KG in the GODB. The fundamentals for the extraction and generation prompt can be summarized as follows: $(1)$ \textbf{Role}: This sets a statement of the role the LLM agent should play, for example, \textit{"You are an assistant in the data science team of a construction materials company aiming to extract information from the minutes of meetings between its sales representatives and clients, with the goal of capturing that knowledge in knowledge graphs in Neo4j"}. $(2)$ \textbf{Instruction}: This provides specific instructions to build the KG, such as \textit{"Could you construct a knowledge graph considering the following premises? (a) It's important to break down the data that can be extracted from the objectives, meeting summaries, and topics discussed. (b) [...]"}. $(3)$ \textbf{Output format}:  It's crucial to define the specific output needed for the data pipeline processing workflow, for instance, \textit{"Format the output only with the Cypher statements necessary to create the graph, considering that several nodes may already exist in the graph"}. In our experiments, we verified that this zero-shot/few-shot approach was able to achieve high performance, with error rates ranging from 3\% to 4\% in Cypher output. Although these results seem to be satisfactory, AI engineers should consider several areas for improvement:
\subsection{Agentic Workflow}
One common issue in this process is the assumption that LLMs are capable of consistently producing excellent results in zero-shot mode. Similar to human reasoning, it's often necessary to iterate repeatedly to generate valuable insights. Fortunately, there is a growing trend to approach these solutions in an Agentic Workflow style \cite{narayana}, which outperforms any zero-shot approach by far. It's important to keep in mind that many iterations for knowledge extraction and KG building will likely be needed, each potentially requiring a different approach or focus. AI engineers should carefully design this task.
\subsection{KG Instruction-tuning}
The instruction section of the prompts is crucial to ensuring success in this process. It's not easy to define these instructions \textit{a priori}. Both domain knowledge and data analysis are required to accurately define this set of instructions. For example, in KG, it's crucial to label relations accurately. The language used in our case may vary from person to person. One important approach is to normalize relation names as early as possible. While it's possible to normalize them in another stage of the pipeline, the sooner this is done, the better. One common approach is the \textbf{exploratory approach}, where AI engineers run the extraction with production data and shape the KG accordingly. Shaping the KG involves normalizing entity and relation names, standardizing and unifying units (such as money, weight, distance, etc.), and refining the set of instructions for improved performance. It's important to find the right \textbf{trade-off} between a \textbf{static graph} schema and a \textbf{free-form} one. If a static graph is defined, there may be no difference between using a GODB or another type of database. However, a completely free and denormalized graph schema may result in low performance or pose a risk of hallucinations.
\subsection{Instruction meta-data}
During KG instruction-tuning, the refined set of instructions may often expand significantly beyond token limits for a prompt. Maintaining a large instruction set within the prompt can result in issues during LLM execution, such as high costs, low accuracy, and potential hallucinations. In such cases, we recommend creating an \textbf{instruction KG} that embeds the necessary knowledge for feature extraction from documents and the creation of the data KG. A Retrieval-Augmented Generation (RAG)\cite{rag} process is needed to be defined over instruction KG in order to let the LLM follow instructions on a particular document. needs to be defined over the instruction KG to enable the LLM to follow instructions on a particular document. The emergence of frameworks like \textit{langchain} \cite{langchain} has simplified RAG processes as well as GODB integration.
\subsection{Quality metrics for the ingestion process} The framework we've established for the ingestion process allows for the incorporation of \textbf{quality metrics for KG construction}. Once AI engineers define a specific use case, the KG can undergo validation with subject matter experts (SMEs), enabling the creation of automated test suites comprising input documents and KG outputs. These test suites facilitate regression testing, ensuring the preservation of functionality over time. Similar to quality assurance in software engineering, it's imperative to maintain these test suites consistently. GODB supports AI engineers in identifying discrepancies in the generated graphs when matches aren't perfect. Additionally, engineers should recognize that LLMs excel at explaining these graph differences and their semantics.

\section{Retrieval-Augmented Generation with KG and GODB}
Once a KG on data is created, numerous advanced analytics paths become available: we can extract insights directly from the GODB itself, or we can construct intelligent agents with LLMs to leverage the KG, as in our case. The importance of applying RAG on GODB instead of building vector embeddings of minutes (in our case) is based on the fact that many user questions can result in a large number of vector matches after running cosine vector similarity match (or any other vector matching function). Just imagine questions like \textit{"give me the volume of cement or concrete sales lost due to humidity issues in 2023"}: probably, a large number of vectors can emerge as candidates, so the standard solution is not suitable for our case. The Langchain framework has introduced a highly maintainable approach to working with RAG \cite{bratanic}, with a focus on the consumption side. We will explore three possible patterns:

\subsection{RAG with GODB} 
Prompt engineering patterns can be implemented to answer user questions using KG in GODB. The Langchain framework can assist AI engineers in implementing a chain of prompts (via chain, agent, or tooling) where the user prompt can be split into short pieces, each of which can be translated to Cypher. Each query can then be executed in Neo4j, and finally, the LLM can interpret the query responses, combine them, and elaborate a conclusive answer \cite{bratanic}. A good use case for this pattern consists of numerical questions, grouping logic, etc. These types of questions can be accurately answered from the perspective of a graph rather than through the elaboration of large amounts of text. Obviously, this approach outperforms RAG over plain text embeddings.

\subsection{RAG with vector index and GODB} 
This approach, referred to as Hybrid Retrieval for RAG \cite{bratanic}, combines the benefits of RAG with GODB and classic vector index embeddings. In this case, the langchain agent can query both the GODB and the vector database after analyzing the user prompt. Then it collects all results, and the LLM is capable of producing a text answer for the user. This approach is a very elegant solution for an agent that needs to deal with questions that can be solved easily with GODB support, but also can take advantage of text search. A good example of a use case for this approach can be the following question: \textit{"give me the volume of cement or concrete sales lost due to humidity issues in 2023 in adherence to law 13943"}. Here, the agent needs to discover the implications of law 13943 and then calculate the sales lost.
\subsection{RAG with embedded vectors in GODB}
The most advanced technique is to integrate embeddings (vectors) into the graph. Neo4j supports vector embeddings in the graph, enabling not only similarity search but also graph operations. Chunks (vectors) become a specific type of nodes connected to the extended KG. Therefore, vector filtering is powered by graph operations, which may traverse paths inside the graph or compute complex algorithms using the GODB's high-performance engine to determine which vectors better match the user prompt, outperforming standard cosine similarity search. The possibilities of this approach extend beyond the scope of this work and represent an open research field.

\section{Enhanced Explainability}
LLMs have introduced an issue regarding explainability. Machine Learning explainability traditionally addresses non-black box logic. However, complex neural networks have introduced behaviors that are more difficult to explain. One potential approach to address explainability in LLMs is through the prompt engineering pattern called Chain of Thought (CoT). It constitutes a prompt engineering technique aimed at enhancing the reasoning capabilities of large language models (LLMs) by decomposing complex problems into intermediate steps. This strategy enables the model to allocate increased computational resources to each step, thereby enhancing the accuracy and efficiency of problem-solving.

Knowledge graphs serve as structured representations of facts and their interconnections, allowing AI systems to retrieve information by traversing the graph to locate relevant nodes (entities or concepts) and edges (relationships) corresponding to the query. This approach offers a subset of the graph as a foundation for fact-checking in LLMs, rendering the process both explainable and grounded, and incorporating hierarchical relationships. GODB play a pivotal role in enabling this capability.

\section{Memory and Context}
In LLM solutions, memory is only introduced when implementing a chatbot. In such instances, the typical approach involves incorporating memory context into each LLM interaction, which includes previous questions and answers. The LLM service remains entirely stateless. Typically, when the prompt size exceeds the token limit, memory is paginated, resulting in the loss of the initial interaction. More sophisticated patterns involve summarizing the chat history and integrating it into the prompt.

GODB can enhance these chatbots' memory representation. Consider creating a graph containing information about chat history and updating it with each interaction. A notable aspect of this approach is that old facts can be replaced by new ones at any given moment. This implementation resembles a session-oriented KG containing user interaction data, which may eventually transition from session-based to user history-based.

This capability opens up the possibility of implementing customer personalization. By leveraging history-based memory, we can extract features that characterize individual customers and construct a personalized KG with this information. LLMs can then utilize this personalized KG through RAG to make decisions that align with customer preferences. An extension of this approach involves connecting similar profiles, enabling LLMs to make recommendations for users with similar profiles. GODB serves as an excellent enabler for implementing these kinds of extensions.

\section{Factuality for avoiding Hallucinations}
Hallucinations pose a significant challenge for LLM applications, and considerable efforts are underway in industry-specific domains \cite{sohini}. A useful initial step in assessing the risk of hallucinations is to evaluate the factual adherence of our application's responses. When utilizing KGs supported by GODB, leveraging GODB to assess factual adherence is advisable.

While this approach is effective for analytical purposes, it may incur a latency penalty if performed online. Consider a model output $R$, which is then split into chunks requiring verification with the assistance of an LLM. We have a sequence of short sentences ${R_1, R_2, .., R_n}$. For each $R_i$ sentence, factual adherence can be checked against other $R_k$ chunks (where $k<i$) with the help of an LLM. The remaining $R$ sentences can be analyzed against our KG with LLM assistance, which may provide support for these statements. Although this process requires exceptions to be added to the base metric model, it enables the calculation of a factual-adherence metric, serving as a Hallucination Risk indicator. Moreover, this process can be easily automated.

\section{Conclusion and future directions}
This study has highlighted the significance of the collaboration between GODB and LLMs in developing domain-specific solutions for industry-specific applications. Drawing on our consulting experience in this field, we emphasize the importance of considering the partnership between GODB and LLMs as a primary approach for implementing Generative AI solutions.

Moving forward, our research agenda will focus on metrics exploration. By leveraging graph theory, we aim to identify and utilize potential graph-based indicators that can enhance the evaluation metrics for LLM applications.
%
%
%

\begin{thebibliography}{10}

\bibitem{accen}
Accenture (Eds.) In: Accenture Blogs, \url{https://www.accenture.com/content/dam/ accenture/final/accenture-com/document/Accenture-A-New-Era-of-Generative-AI-for-Everyone.pdf} [March 2024], Accenture (2023).

\bibitem{graphdb}
Angles, R.: A Comparison of Current Graph Database Models. In proceedings: 2012 IEEE 28th International Conference on Data Engineering Workshops, doi: 10.1109/ICDEW.2012.31., pp. 171-177, Arlington, VA, USA (2012).

\bibitem{langchain}
Auffarth, B.: Generative AI with LangChain: Build large language model (LLM) apps with Python, ChatGPT and other LLMsGenerative AI with LangChain: Build large language model (LLM) apps with Python, ChatGPT and other LLMs. ISBN 978-1835083468. Pakt Publishing. USA (2023).

\bibitem{bechberger}
Bechberger D, Perryman, J.: Exploring Graph Databases. Manning Publications Co. ISBN: 9781617299674
(2021)

\bibitem{bordes}
Bordes, A., Weston, J., Collobert, R., Bengio, Y.: Learning Structured Embeddings of Knowledge Bases. In Proceedings: Twenty-Fifth AAAI Conference on Artificial Intelligence, AAAI 2011, San Francisco, California, USA, August 7-11, 2011.

\bibitem{bratanic}
Bratanic, T.: Enhancing the Accuracy of RAG Applications With Knowledge Graphs. In: Medium. 
\url{https://medium.com/neo4j/enhancing-the-accuracy-of-rag-applications-with-knowledge-graphs-ad5e2ffab663} [April, 2024], Medium (2024).

\bibitem{ehrlinger}
Ehrlinger, L., Wöß, W.: Towards a Definition of Knowledge Graphs. In Joint Proceedings of the Posters and Demos Track of 12th International Conference on Semantic Systems - SEMANTiCS 2016 and 1st International Workshop on Semantic Change and Evolving Semantics (SuCCESS16). Volume: 1695. Leipzig, Germany (2016).

\bibitem{farber}
Farber, M., Ell, B.. Menne, C., Rettinger, A., Bartscherer, F. Linked Data Quality of DBpedia, Freebase, OpenCyc, Wikidata, and YAGO. In: Semantic Web Journal. \url{http://www.semantic-web-journal.net/content/linked-data-quality-dbpedia-freebaseopencyc-wikidata-and-yago} [March, 2024] (2016).

\bibitem{jovanovic}
Jovanovic, M., Campbell, M.: Connecting AI: Merging Large Language Models and Knowledge Graph. Computer. 56. 10.1109/MC.2023.3305206. IEEE (2023).

\bibitem{rag}
Lewis, P., Perez, E., Piktus, A., et al.: In: Advances in Neural Information Processing Systems. v33, pp.9459-9474, Retrieval-Augmented Generation for Knowledge-Intensive NLP Tasks. Curran Associates, Inc. (2020).

\bibitem{narayana}
Narayana, L.: Exploring Agentic Workflows in AI: A Practical Approach with CrewAI, OpeRouter.ai and OpenHermes. In: Medium. \url{https://blog.stackademic.com/exploring-agentic-workflows-in-ai-a-practical-approach-with-crewai-operouter-ai-and-openhermes-cb7abd493285} [April 2024] (2024).

\bibitem{pan}
Pan, S., Linhao, L., Wang, Y., Chen, C., Wang J., Wu, X.: Unifying Large Language Models and Knowledge Graphs: A Roadmap. Preprint. \url{https://api.semanticscholar.org/CorpusID:259165563}, abs/2306.08302. (2023).
  
\bibitem{peng}
Peng, C., Xia, F., Naseriparsa, M, Osborne, F.: Knowledge Graphs: Opportunities and Challenges. In: Artificial Intelligence Review. 56. 1-32. 10.1007/s10462-023-10465-9. Springer (2023). 

\bibitem{google}
Singhal, A.: Introducing the Knowledge Graph: Things, not Strings. In: Google Blogs. \url{https://googleblog.blogspot.co.at/2012/05/introducing-knowledge-graph-things-not} [March, 2024], Google (2012) 

\bibitem{sohini}
Roychowdhury, S., Alvarez, A., Moore, B., Krema, M., Gelpi, M.P., Rodríguez, F.M., Cabrejas, J.R., Martínez Serrano, P., Agrawal, P., Mkherjee, A.: Hallucination-minimized Data-to-answer Framework for Financial Decision-makers. Preprint. \url{https://arxiv.org/pdf/2311.07592.pdf} [March 2024] (2023).


\end{thebibliography}
%

\end{document}